\documentclass{article}
\usepackage{spconf,amsmath,amssymb,graphicx,hyperref}

\title{LCAP: POPULATION-INFORMED LATENT CHIP ADAPTATION FROM FEW OUTPUT PROBES FOR PHOTONIC NEURAL NETWORKS}

\name{Tianyu Gao$^{1,2,*}$ and Guantian Zheng$^{3}$}

\address{
$^{1}$City University of Hong Kong, Hong Kong SAR, China;
$^{2}$Sichuan University, Chengdu, China\\
$^{3}$Nanyang Technological University, Singapore;
$^{*}$gaotianyu@stu.scu.edu.cn
}

\begin{document}
\maketitle

\begin{abstract}
Photonic neural networks (PNNs) offer efficient analog inference, but parameters optimized under ideal device models can degrade after fabrication, creating a persistent simulation-to-hardware (sim-to-real) gap. When many identically designed chips are deployed, calibrating each device from scratch compounds this cost. We propose Latent Chip Adaptation from Probes (LCAP), a population-informed framework that decomposes hardware adaptation into a transferable population correction and probe-inferred latent personalization. LCAP first learns a shared correction from 80 historical chips, then extracts a low-dimensional correction space from device-specific refinements. At deployment, 32 fixed unlabeled output probes infer an unseen chip's latent correction coordinates, enabling feed-forward personalization without target-device optimization. On a three-layer 64-mode MZI simulator with phase variation, beam-splitter errors, quantization, and crosstalk, accuracy improves from 80.4147\% under direct deployment to 92.6860\% after shared calibration and 93.3617\% with LCAP. LCAP improves 27/30 unseen chips and raises worst-device accuracy from 89.18\% to 90.54\%.
\end{abstract}


\section{Introduction}
\label{sec:intro}

\begin{figure}[t]
    \centering
    \includegraphics[width=\columnwidth]{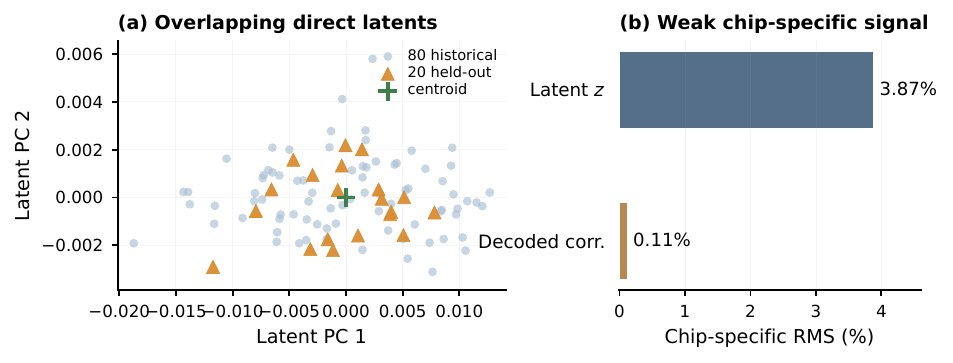}
    \caption{Motivating diagnostic for direct latent adaptation. Relative chip-specific RMS measures the deviation from the population mean normalized by the total RMS. In an early encoder--decoder prototype, chip-specific variation accounts for only 3.87\% of the latent RMS and 0.11\% of the decoded-correction RMS, suggesting that population-common structure can dominate direct end-to-end representations.}
    \label{fig:latent_teaser}
\end{figure}

Photonic neural networks (PNNs) implement linear transformations directly in optical interference meshes, offering a promising route toward high-throughput and energy-efficient inference \cite{Shen2017,Clements2016,Shastri2021,Bogaerts2020}. Their performance, however, is tightly coupled to physical device parameters. Parameters optimized with an ideal transfer model are ultimately executed on fabricated circuits. Their phase shifts, splitting ratios, and control responses can deviate from design, while quantization and crosstalk introduce additional mismatch. When compounded through MZI meshes, these nonidealities can substantially degrade task accuracy \cite{Banerjee2023}. More importantly, the resulting sim-to-real gap is not identical across devices: chips fabricated from the same design can exhibit different persistent deviations and therefore require different corrections \cite{Xing2023}. As deployment scales from a single prototype to many chips, independently recalibrating each device turns adaptation into a recurring cost. A natural question follows: \emph{can calibration experience accumulated from earlier chips be reused to adapt a new chip?}

Existing approaches mainly address this deployment gap through two paradigms. Device-specific methods, including L$^2$ight, DAT, and meta-learning-based on-chip training, reduce the cost of adapting each physical system but still require optimization tied to the target device \cite{Hughes2018,Gu2021L2ight,Zheng2023DAT,Ho2025Meta}. Transfer-oriented methods instead seek a common solution that remains robust across hardware instances, as exemplified by Transferable Learning and sharpness-aware training (SAT) \cite{Vadlamani2023Transferable,Xu2026SAT}. Between these paradigms lies a less explored possibility: reusing population-level calibration experience while still preserving the individuality of each new chip. A natural attempt is to directly encode probe responses into a hardware
latent and decode them into full device corrections. However, an early
direct probe-to-correction encoder--decoder produced strongly overlapping
chip latents across both historical and held-out devices
(Fig.~\ref{fig:latent_teaser}a). Moreover, the chip-specific component
accounted for only 3.87\% of the latent RMS and only 0.11\% of the
decoded-correction RMS (Fig.~\ref{fig:latent_teaser}b). This points to an
identifiability problem: population-common mismatch can dominate an
end-to-end representation and obscure the weaker device-specific structure.

\begin{figure*}[t]
    \centering
    \includegraphics[width=\textwidth]{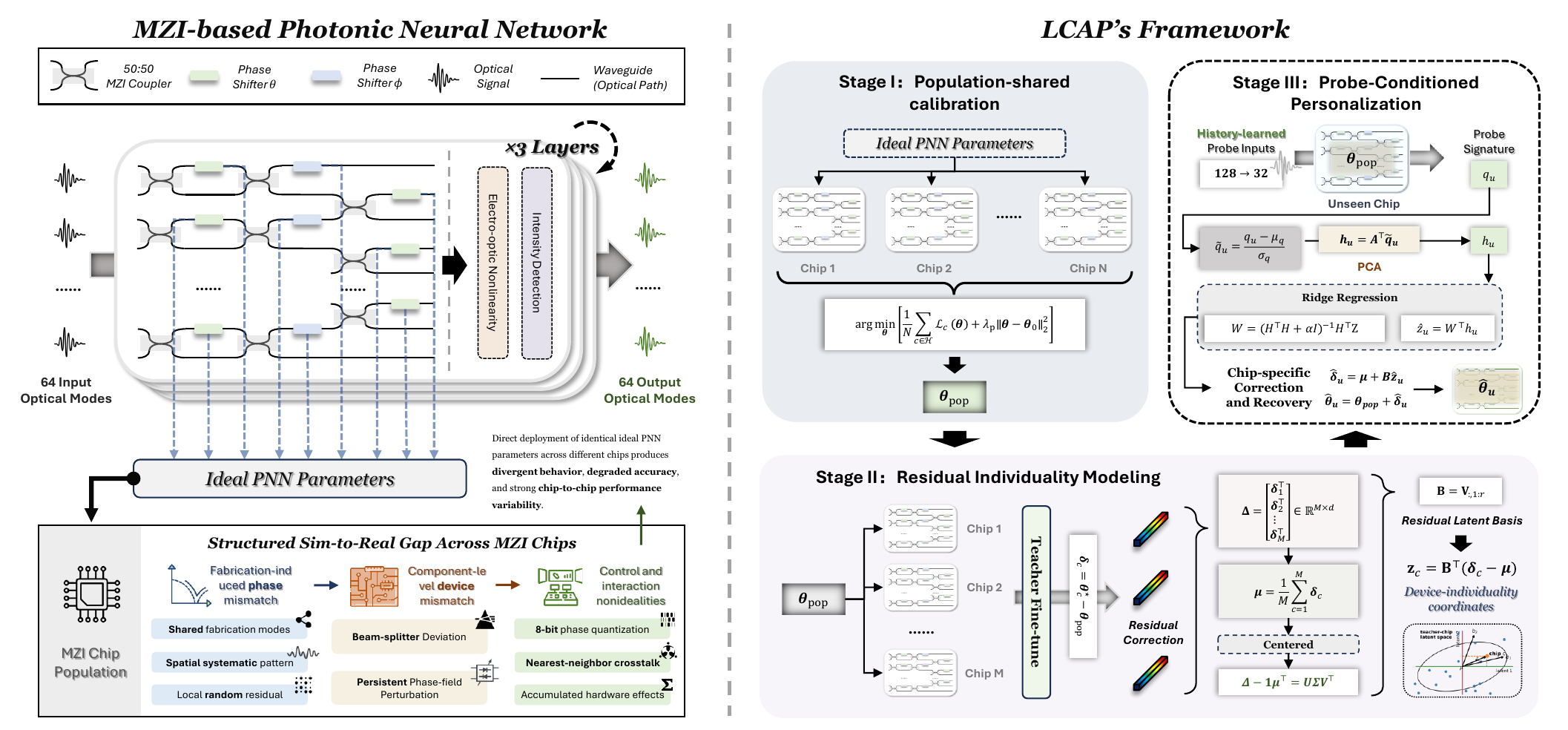}
    \caption{LCAP learns a population anchor, models residual device individuality, and infers an unseen chip's correction from fixed probes without target-device optimization.}
    \label{fig:framework}
\end{figure*}

Motivated by this observation, we propose Latent Chip Adaptation from Probes (LCAP), a common-first, individuality-second framework that first learns a transferable population calibration and then models only the remaining device-specific corrections in a low-dimensional latent space (Fig.~\ref{fig:framework}). For an unseen chip, fixed unlabeled probes infer its latent coordinates and reconstruct a personalized correction without target-device optimization. Across 30 unseen chips, LCAP improves the deployment trajectory from 80.41\% to 92.69\% and finally 93.36\%.

\section{Method: Latent Chip Adaptation from Probes}
\label{sec:method}

\noindent\textbf{Overview.}
Let $\boldsymbol{\theta}_0\in\mathbb{R}^{d}$ denote the programmable
parameters of a source PNN optimized under the ideal device model, and let
$f_c^{\mathrm{hw}}(\cdot;\boldsymbol{\theta})$ denote its realization on
chip $c$. LCAP follows a \emph{common-first, individuality-second}
principle (Fig.~\ref{fig:framework}). For an unseen chip $u$, its
adapted parameters are written as
\begin{equation}
\hat{\boldsymbol{\theta}}_u
=
\boldsymbol{\theta}_{\mathrm{pop}}
+
\hat{\boldsymbol{\delta}}_u ,
\label{eq:lcap_decomp}
\end{equation}
where $\boldsymbol{\theta}_{\mathrm{pop}}$ captures adaptation transferable
across a historical chip population, while
$\hat{\boldsymbol{\delta}}_u$ models the remaining device-specific
correction.

\noindent\textbf{Population-shared calibration.}
Given $N$ historical chips $\mathcal{H}$, we optimize one common parameter
vector over the entire population:
\begin{equation}
\boldsymbol{\theta}_{\mathrm{pop}}
=
\arg\min_{\boldsymbol{\theta}}
\left[
\frac{1}{N}
\sum_{c\in\mathcal{H}}
\mathcal{L}_{c}(\boldsymbol{\theta})
+
\lambda_{\mathrm p}
\|\boldsymbol{\theta}-\boldsymbol{\theta}_0\|_2^2
\right],
\label{eq:population}
\end{equation}
where
\begin{equation}
\mathcal{L}_{c}(\boldsymbol{\theta})
=
\mathbb{E}_{(\mathbf{x},y)}
\left[
\ell\!\left(
f_c^{\mathrm{hw}}(\mathbf{x};\boldsymbol{\theta}),y
\right)
\right].
\label{eq:chip_loss}
\end{equation}
This stage does not impose a low-dimensional constraint. Instead, it learns
a full shared PNN configuration that captures the transferable component
of population-level hardware mismatch before chip individuality is
modeled.

\noindent\textbf{Residual latent device space.}
For $M$ historical teacher chips $\mathcal{T}\subset\mathcal{H}$, we further
optimize each device starting from $\boldsymbol{\theta}_{\mathrm{pop}}$,
yielding device-specific teacher parameters
$\boldsymbol{\theta}_{c}^{\star}$. We define the residual correction as
\begin{equation}
\boldsymbol{\delta}_{c}
=
\boldsymbol{\theta}_{c}^{\star}
-
\boldsymbol{\theta}_{\mathrm{pop}} .
\label{eq:residual}
\end{equation}
Stacking all teacher residuals into
$\mathbf{\Delta}\in\mathbb{R}^{M\times d}$, we center the matrix and apply
SVD:
\begin{equation}
\mathbf{\Delta}
-
\mathbf{1}\boldsymbol{\mu}^{\top}
=
\mathbf{U}\mathbf{\Sigma}\mathbf{V}^{\top}.
\label{eq:svd}
\end{equation}
The first $r$ right-singular vectors form the latent correction basis
\begin{equation}
\mathbf{B}
=
\mathbf{V}_{:,1:r},
\qquad
\mathbf{z}_{c}
=
\mathbf{B}^{\top}
(\boldsymbol{\delta}_{c}-\boldsymbol{\mu}),
\label{eq:latent_coord}
\end{equation}
such that
\begin{equation}
\boldsymbol{\delta}_{c}
\approx
\boldsymbol{\mu}
+
\mathbf{B}\mathbf{z}_{c}.
\label{eq:latent_recon}
\end{equation}
Here, $\mathbf{z}_{c}\in\mathbb{R}^{r}$ represents the coordinates of
device individuality in the residual correction space. Importantly, LCAP
does not compress the full PNN parameters; it models only what remains
device-specific after population calibration.

\noindent\textbf{Probe-conditioned latent inference.}
At deployment, the teacher refinement above is unavailable for an unseen
chip. LCAP therefore infers its latent coordinates from a small set of
hardware measurements. For a candidate probe $\mathbf{p}$, we define the
output residual
\begin{equation}
\begin{aligned}
\mathbf{r}_{c}(\mathbf{p})
=
\mathrm{Re}\big[
&f_c^{\mathrm{hw}}
(\mathbf{p};\boldsymbol{\theta}_{\mathrm{pop}})
\\[-1mm]
&-
f^{\mathrm{id}}
(\mathbf{p};\boldsymbol{\theta}_{\mathrm{pop}})
\big].
\end{aligned}
\label{eq:probe_residual}
\end{equation}
A fixed probe set $\mathcal{P}^{\star}$ is selected offline using historical
chips by ranking candidate probes according to their cross-chip response
variance. Concatenating the selected probe residuals gives an observable
hardware signature $\mathbf{q}_{c}$.

We standardize $\mathbf{q}_{c}$ and project it to a compact probe feature
$\mathbf{h}_{c}$ using PCA:
\begin{equation}
\mathbf{h}_{c}
=
\mathbf{A}^{\top}
\tilde{\mathbf{q}}_{c}.
\label{eq:probe_feature}
\end{equation}
Given historical feature and latent matrices $\mathbf{H}$ and $\mathbf{Z}$,
ridge regression learns the probe-to-latent mapping
\begin{equation}
\mathbf{W}
=
\left(
\mathbf{H}^{\top}\mathbf{H}
+
\alpha\mathbf{I}
\right)^{-1}
\mathbf{H}^{\top}\mathbf{Z}.
\label{eq:ridge}
\end{equation}
For an unseen chip $u$,
\begin{equation}
\hat{\mathbf{z}}_{u}
=
\mathbf{W}^{\top}\mathbf{h}_{u},
\label{eq:infer_z}
\end{equation}
and its residual correction is reconstructed as
\begin{equation}
\hat{\boldsymbol{\delta}}_{u}
=
\boldsymbol{\mu}
+
\mathbf{B}\hat{\mathbf{z}}_{u}.
\label{eq:infer_delta}
\end{equation}
The final deployed parameters are therefore
\begin{equation}
\hat{\boldsymbol{\theta}}_{u}
=
\boldsymbol{\theta}_{\mathrm{pop}}
+
\hat{\boldsymbol{\delta}}_{u}.
\label{eq:final_deploy}
\end{equation}
Thus, once the population model and probe-to-latent mapping are learned
offline, a new chip requires only a fixed set of unlabeled output
measurements. The remaining high-dimensional adaptation problem is reduced
to low-dimensional latent inference, without target-device optimization.

\section{Experiments}
\label{sec:experiments}

\subsection{Experimental Setup}

\noindent\textbf{PNN and task.}
We follow the MZI-PNN simulation configuration used in DAT
\cite{Zheng2023DAT}: a three-layer, 64-mode MZI network for MNIST
classification. Each image is Fourier transformed, and its central
$8\times8$ spectrum forms the 64-dimensional complex optical input. The
same ideal PNN parameters initialize all virtual hardware instances.

\noindent\textbf{Chip population.}
DAT and SAT model persistent phase-shifter and beam-splitter deviations as
MZI fabrication errors \cite{Zheng2023DAT,Xu2026SAT}. Building on this
component-level model, we generate a structured chip population with
\begin{equation}
\begin{aligned}
\Delta\boldsymbol{\phi}_c
&=\sigma_{\phi,c}\,
{\rm Norm}\!\left(
\sqrt{.50}\,\mathbf{s}_c+
\sqrt{.25}\,\mathbf{g}+
\sqrt{.25}\,\boldsymbol{\epsilon}_c
\right),\\[-1mm]
\mathbf{s}_c
&={\rm Norm}\!\left(
\sum_{k=1}^{8}a_{c,k}\mathbf{B}_k
\right),\quad
a_{c,k}\sim\mathcal{N}(0,1).
\end{aligned}
\label{eq:chip_error}
\end{equation}
Here, $\{\mathbf{B}_k\}$ are shared fabrication modes with independently
sampled chip-specific coefficients, $\mathbf{g}$ is a fixed low-frequency
spatial systematic pattern, and $\boldsymbol{\epsilon}_c$ is an independent
local residual; all components are RMS-normalized before mixing. This
shared--spatial--local construction reflects the coexistence of correlated,
spatially varying, and local process variations reported in silicon photonics
\cite{Chen2013Process,Xing2023,Boning2022,Suda2026}. The 50/25/25 mixture
and eight shared modes are modeling choices rather than assumptions about
the intrinsic physical rank of fabrication variation. We use
$\sigma_{\phi,c}\!\sim\!\mathcal{U}(0.03,0.07)$ and independent
beam-splitter errors $\sigma_{bs,c}\!\sim\!\mathcal{U}(0.02,0.04)$
\cite{Zheng2023DAT,Xu2026SAT}. Each error realization remains fixed for a
chip. We additionally apply 8-bit phase quantization and nearest-neighbor
crosstalk ($0.005$), following L$^2$ight \cite{Gu2021L2ight}; such combined
nonidealities are known to accumulate in coherent MZI meshes
\cite{Banerjee2023}.

\noindent\textbf{Protocol.}
We use 80 historical chips for population calibration, including 40 teacher
chips refined for 300 steps. All probe selection and latent/regression
fitting use historical devices only. The frozen LCAP configuration uses
rank $r=16$, 16 PCA components, Ridge $\alpha=100$, and 32 probes selected
from 128 fixed $\{-1,+1\}^{64}$ candidates by historical cross-chip response
variance. Final results are reported on 30 independently generated, severity-matched
chips, each evaluated on all 10,000 MNIST test images. These test chips are
resampled independently from the same population model and severity ranges,
and never participate in probe design, latent fitting, or model selection.

\subsection{Hierarchical Sim-to-Real Recovery}

Table~\ref{tab:main} reports the hierarchical deployment trajectory on the
same 30 unseen chips. Directly transferring the ideal PNN yields only
80.4147\% mean accuracy, with a 9.19-point cross-chip standard deviation and
a worst-device accuracy of 60.21\%, confirming substantial device-dependent
sim-to-real mismatch. Population-shared calibration recovers 12.2713 points,
raising the mean to 92.6860\%. All 30 chips improve, the cross-chip standard
deviation contracts to 1.27 points, and the worst-device accuracy rises to
89.18\%, indicating that the dominant deployment loss is highly transferable.

LCAP then targets the residual individuality left by this strong shared
solution. With 32 unlabeled probes and no target-device optimization, it
reaches 93.3617\% ($+0.6757$ pt), improves 27/30 unseen chips, and raises the
worst-device accuracy to 90.54\%. The
$80.41\!\rightarrow\!92.69\!\rightarrow\!93.36$ progression supports the
common-first, individuality-second design: most mismatch is shared, while
the remaining device-specific correction is still predictable from sparse
hardware observations.

\begin{table}[t]
\centering
\caption{Hierarchical deployment recovery on the same 30 unseen chips.
LCAP uses no target-chip labels or target-device optimization.}
\label{tab:main}
\small
\setlength{\tabcolsep}{4.5pt}
\begin{tabular}{lccc}
\hline
Metric
& Direct
& Population
& \textbf{LCAP} \\
\hline
Mean accuracy (\%)
& 80.415
& 92.686
& \textbf{93.362} \\

Gain over previous (pt)
& --
& +12.271
& \textbf{+0.676} \\

Worst-device acc. (\%)
& 60.21
& 89.18
& \textbf{90.54} \\

Across-chip std. (pt)
& 9.189
& 1.274
& \textbf{1.198} \\

Improved chips
& --
& 30/30
& \textbf{27/30} \\

New-chip adaptation
& None
& None
& \textbf{32 probes} \\
\hline
\end{tabular}
\end{table}

\subsection{Latent Compactness and Probe Observability}

LCAP relies on two properties of the residual correction space: device
individuality should be compact enough to admit a low-dimensional
representation, yet observable enough to be inferred from sparse hardware
responses. Table~\ref{tab:ablation} examines both properties while keeping
the remaining deployment protocol fixed.

\noindent\textbf{Latent dimensionality.}
Increasing the rank from 8 to 32 raises captured teacher-correction energy
from 43.1\% to 91.4\%. However, unseen-chip accuracy is not monotonic:
$r=16$ gives the highest mean accuracy of 93.3617\%, whereas $r=32$ falls
slightly to 93.3160\% despite retaining substantially more teacher energy.
Thus, deployment utility depends not only on reconstruction capacity, but
also on whether the retained correction directions can be reliably inferred
for an unseen device.

\noindent\textbf{Probe observability.}
With $r=16$ fixed, increasing the probe budget from 8 to 16 and 32 improves
mean accuracy from 93.0827\% to 93.2497\% and 93.3617\%, respectively. The
gain over population calibration increases from $+0.3967$ to $+0.5637$ and
$+0.6757$ pt, while the number of improved chips rises from 22/30 to 27/30.
These results favor a compact latent whose coordinates remain identifiable
from sparse hardware observations.

\begin{table}[t]
\centering
\caption{Rank and probe-count ablations on the same 30 unseen chips.}
\label{tab:ablation}
\small
\setlength{\tabcolsep}{4.2pt}

\begin{tabular}{ccccc}
\hline
\multicolumn{5}{c}{\textbf{(a) Residual latent rank}} \\
\hline
Rank
& Energy (\%)
& Mean (\%)
& Gain (pt)
& Improved \\
\hline
8
& 43.1
& 93.3080
& +0.6220
& 27/30 \\

\textbf{16}
& \textbf{63.5}
& \textbf{93.3617}
& \textbf{+0.6757}
& 27/30 \\

32
& 91.4
& 93.3160
& +0.6300
& 28/30 \\
\hline

\multicolumn{5}{c}{}\\[-1.5mm]
\multicolumn{5}{c}{\textbf{(b) Number of output probes}} \\
\hline
Probes
& --
& Mean (\%)
& Gain (pt)
& Improved \\
\hline
8
& --
& 93.0827
& +0.3967
& 22/30 \\

16
& --
& 93.2497
& +0.5637
& 27/30 \\

\textbf{32}
& --
& \textbf{93.3617}
& \textbf{+0.6757}
& \textbf{27/30} \\
\hline
\end{tabular}
\end{table}

\begin{figure}[t]
    \centering
    \includegraphics[width=\columnwidth]
    {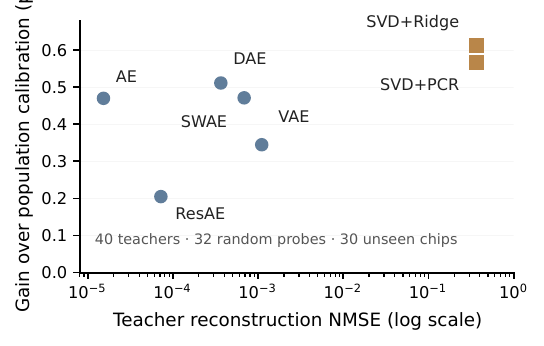}
\caption{
\textbf{Reconstruction fidelity does not imply deployment utility.}
Under the same 40-teacher, 32-probe, 30-chip protocol, autoencoder
representations reconstruct historical corrections more accurately, whereas
the rank-16 linear residual space yields larger unseen-chip adaptation gains.
}
    \label{fig:representation}
\end{figure}

\subsection{Reconstruction vs.\ Deployment Utility}

Figure~\ref{fig:representation} compares residual representations under the
same 40-teacher, 32-random-probe protocol. Autoencoder variants reconstruct
historical teacher corrections extremely accurately, yet this fidelity does
not translate into better deployment. Most strikingly, AE achieves a teacher
reconstruction NMSE of only $1.52\times10^{-5}$, versus 0.365 for rank-16
SVD, but its unseen-chip gain is smaller ($+0.470$ vs.\ $+0.612$ pt with
SVD+Ridge). DAE, SWAE, and VAE show the same general mismatch.

This inversion highlights a key distinction between compression and
adaptation: a latent may preserve teacher-specific details that are difficult
to infer from a small output signature. In contrast, the linear residual
space can discard such details while retaining directions that remain
predictable across devices. Consistent with this interpretation, PLS reaches
$+0.636$ pt without explicit reconstruction, and replacing random probes with
historically selected informative probes raises the frozen LCAP configuration
to 93.3617\%. A deployable hardware latent should therefore be judged by
predictability on unseen chips, not reconstruction fidelity alone.

\section{Conclusion}

We introduced LCAP, a population-informed framework that decomposes PNN
hardware adaptation into population-shared calibration and probe-conditioned
residual personalization. Across 30 unseen MZI chips, shared calibration
recovers direct-deployment accuracy from 80.41\% to 92.69\%, while 32
unlabeled output probes further raise it to 93.36\% without target-device
optimization. Rank, probe-count, and representation ablations show that a
useful hardware latent should be compact and identifiable from sparse
observations rather than merely reconstructive. These results support a
\emph{common-first, individuality-second} paradigm for scalable sim-to-real
adaptation of photonic neural networks.

\section{Compliance with Ethical Standards}
This is a numerical simulation study using the publicly available MNIST
benchmark; no ethical approval was required.

\section{Acknowledgment}
No funding was received for conducting this study. The authors have no
relevant financial or nonfinancial interests to disclose.

%

\bibliographystyle{IEEEbib}
\bibliography{refs_LCAP}

@article{Shen2017,
  author  = {Yichen Shen and Nicholas C. Harris and Scott Skirlo and Mihika Prabhu and Tom Baehr-Jones and Michael Hochberg and Xin Sun and Shijie Zhao and Hugo Larochelle and Dirk Englund and Marin Solja{\v{c}}i{\'c}},
  title   = {Deep Learning with Coherent Nanophotonic Circuits},
  journal = {Nature Photonics},
  volume  = {11},
  pages   = {441--446},
  year    = {2017},
  doi     = {10.1038/nphoton.2017.93}
}

@article{Clements2016,
  author  = {William R. Clements and Peter C. Humphreys and Benjamin J. Metcalf and W. Steven Kolthammer and Ian A. Walmsley},
  title   = {Optimal Design for Universal Multiport Interferometers},
  journal = {Optica},
  volume  = {3},
  number  = {12},
  pages   = {1460--1465},
  year    = {2016},
  doi     = {10.1364/OPTICA.3.001460}
}

@article{Shastri2021,
  author  = {Bhavin J. Shastri and Alexander N. Tait and Thomas Ferreira de Lima and Wolfram H. P. Pernice and Harish Bhaskaran and C. David Wright and Paul R. Prucnal},
  title   = {Photonics for Artificial Intelligence and Neuromorphic Computing},
  journal = {Nature Photonics},
  volume  = {15},
  number  = {2},
  pages   = {102--114},
  year    = {2021},
  doi     = {10.1038/s41566-020-00754-y}
}

@article{Banerjee2023,
  author  = {Sanmitra Banerjee and Mahdi Nikdast and Krishnendu Chakrabarty},
  title   = {Characterizing Coherent Integrated Photonic Neural Networks Under Imperfections},
  journal = {Journal of Lightwave Technology},
  volume  = {41},
  number  = {5},
  pages   = {1464--1479},
  year    = {2023},
  doi     = {10.1109/JLT.2022.3193658}
}

@article{Xing2023,
  author  = {Yufei Xing and Jiaxing Dong and Umar Khan and Wim Bogaerts},
  title   = {Capturing the Effects of Spatial Process Variations in Silicon Photonic Circuits},
  journal = {ACS Photonics},
  volume  = {10},
  number  = {4},
  pages   = {928--944},
  year    = {2023},
  doi     = {10.1021/acsphotonics.2c01194}
}

@inproceedings{Gu2021L2ight,
  author    = {Jiaqi Gu and Hanqing Zhu and Chenghao Feng and Zixuan Jiang and Ray Chen and David Z. Pan},
  title     = {L$^2$ight: Enabling On-Chip Learning for Optical Neural Networks via Efficient In-Situ Subspace Optimization},
  booktitle = {Advances in Neural Information Processing Systems},
  volume    = {34},
  pages     = {8649--8661},
  year      = {2021}
}

@article{Zheng2023DAT,
  author  = {Ziyang Zheng and Zhengyang Duan and Hang Chen and Rui Yang and Sheng Gao and Haiou Zhang and Hongkai Xiong and Xing Lin},
  title   = {Dual Adaptive Training of Photonic Neural Networks},
  journal = {Nature Machine Intelligence},
  volume  = {5},
  pages   = {1119--1129},
  year    = {2023},
  doi     = {10.1038/s42256-023-00723-4}
}

@inproceedings{Ho2025Meta,
  author    = {Matthew Ho and Zhanghao Sun and Carson Valdez and Olav Solgaard},
  title     = {Meta-Learning for On-Chip Photonic Neural Network Training},
  booktitle = {CLEO 2025},
  year      = {2025},
  note      = {Paper AA128\_5},
  doi       = {10.1364/CLEO_AT.2025.AA128_5}
}

@article{Vadlamani2023Transferable,
  author  = {Sri Krishna Vadlamani and Dirk Englund and Ryan Hamerly},
  title   = {Transferable Learning on Analog Hardware},
  journal = {Science Advances},
  volume  = {9},
  number  = {28},
  pages   = {eadh3436},
  year    = {2023},
  doi     = {10.1126/sciadv.adh3436}
}

@article{Xu2026SAT,
  author  = {Tengji Xu and Zeyu Luo and Shaojie Liu and Li Fan and Qiarong Xiao and Benshan Wang and Dongliang Wang and Chaoran Huang},
  title   = {Physical Neural Networks Using Sharpness-Aware Training},
  journal = {Nature Communications},
  volume  = {17},
  pages   = {1766},
  year    = {2026},
  doi     = {10.1038/s41467-026-68470-9}
}

@article{Boning2022,
  author  = {Duane S. Boning and Sally I. El-Henawy and Zhengxing Zhang},
  title   = {Variation-Aware Methods and Models for Silicon Photonic Design-for-Manufacturability},
  journal = {Journal of Lightwave Technology},
  volume  = {40},
  number  = {6},
  pages   = {1776--1783},
  year    = {2022},
  doi     = {10.1109/JLT.2021.3115463}
}

@inproceedings{Suda2026,
  author    = {Satoshi Suda and Tadashi Murao and Yuki Atsumi and Ryosuke Matsumoto and Takeru Amano},
  title     = {Die-to-Die Phase-Error Mapping of Silicon MZI Mesh Using Linear-Regression-Assisted Estimation},
  booktitle = {CLEO 2026},
  year      = {2026},
  note      = {Paper JTU.19},
  doi       = {10.1364/CLEO_AT.2026.JTU.19}
}

@article{Hughes2018,
  author  = {Tyler W. Hughes and Momchil Minkov and Yu Shi and Shanhui Fan},
  title   = {Training of Photonic Neural Networks through In Situ Backpropagation and Gradient Measurement},
  journal = {Optica},
  volume  = {5},
  number  = {7},
  pages   = {864--871},
  year    = {2018},
  doi     = {10.1364/OPTICA.5.000864}
}

@article{Bogaerts2020,
  author  = {Wim Bogaerts and Daniel P{\'e}rez and Jos{\'e} Capmany and David A. B. Miller and Joyce Poon and Dirk Englund and Francesco Morichetti and Andrea Melloni},
  title   = {Programmable Photonic Circuits},
  journal = {Nature},
  volume  = {586},
  pages   = {207--216},
  year    = {2020},
  doi     = {10.1038/s41586-020-2764-0}
}

@article{Chen2013Process,
  author  = {Xi Chen and Moustafa Mohamed and Zheng Li and Li Shang and Alan R. Mickelson},
  title   = {Process Variation in Silicon Photonic Devices},
  journal = {Applied Optics},
  volume  = {52},
  number  = {31},
  pages   = {7638--7647},
  year    = {2013},
  doi     = {10.1364/AO.52.007638}
}

\end{document}